\documentclass{article}

\usepackage{microtype}
\usepackage{graphicx}
\usepackage{subfigure}
\usepackage{booktabs} 

\usepackage{hyperref}

\usepackage[accepted]{icml2025}

\usepackage{amsmath}
\usepackage{amssymb}
\usepackage{mathtools}
\usepackage{amsthm}

\usepackage[capitalize,noabbrev]{cleveref}

\theoremstyle{plain}

\theoremstyle{definition}

\theoremstyle{remark}

\usepackage[textsize=tiny]{todonotes}

\icmltitlerunning{Submission for ICML 2025}

\begin{document}

\twocolumn[
\icmltitle{Integration Flow Models}



\icmlsetsymbol{equal}{*}

\begin{icmlauthorlist}
\icmlauthor{Jingjing Wang}{equal,yyy}
\icmlauthor{Dan Zhang}{equal,yyy}
\icmlauthor{Joshua Luo}{yyy}
\icmlauthor{Yin Yang}{sch}
\icmlauthor{Feng Luo}{yyy}
\end{icmlauthorlist}

\icmlaffiliation{yyy}{School of Computing, Clemson University,Clemson, SC, USA}
\icmlaffiliation{sch}{Kahlert School of Computing, University of Utah, USA}

\icmlcorrespondingauthor{Feng Luo}{luofeng@clemson.edu}

\icmlkeywords{Machine Learning, ICML}

\vskip 0.3in
]




\begin{abstract}
Ordinary differential equation (ODE) based generative models have emerged as a powerful approach for producing high-quality samples in many applications. However, the ODE-based methods either suffer the discretization error of numerical solvers of ODE, which restricts the quality of samples when only a few NFEs are used, or struggle with training instability. In this paper, we proposed Integration Flow, which directly learns the integral of ODE-based trajectory paths without solving the ODE functions. Moreover, Integration Flow explicitly incorporates the target state $\mathbf{x}_0$ as the anchor state in guiding the reverse-time dynamics. We have theoretically proven this can contribute to both stability and accuracy. To the best of our knowledge, Integration Flow is the first model with a unified structure to estimate ODE-based generative models and the first to show the exact straightness of 1-Rectified Flow without reflow. Through theoretical analysis and empirical evaluations, we show that Integration Flows achieve improved performance when it is applied to existing ODE-based models, such as diffusion models, Rectified Flows, and PFGM++. Specifically, Integration Flow achieves one-step generation on CIFAR10 with FIDs of 2.86 for the Variance Exploding (VE) diffusion model, 3.36 for rectified flow without reflow, and 2.91 for PFGM++; and on ImageNet with FIDs of 4.09 for VE diffusion model, 4.35 for rectified flow without reflow and 4.15 for PFGM++.
\end{abstract}
\section{Introduction}
\label{sec:intro}

\begin{figure}[ht]
  \centering
  \includegraphics[width=0.47\textwidth]{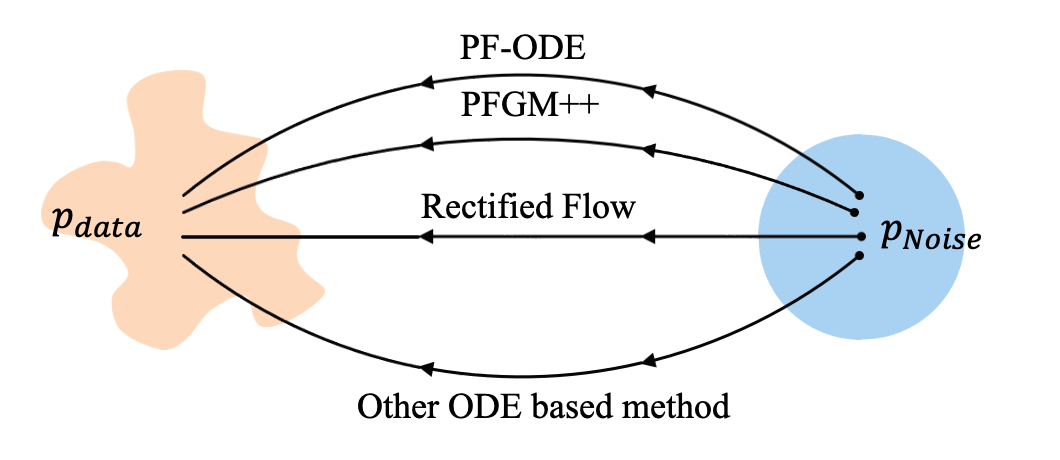}
  \setlength{\abovecaptionskip}{-5pt}
  \caption{An illustration of ODE-based methods, including PF-ODE, PFGM++, and Rectified Flow.}
  \label{fig1}
\end{figure}

Recently, ODE-based generative models have emerged as a cutting-edge method for producing high-quality samples in many applications including image, audio \cite{kong2021diffwave,popov2022diffusionbased}, and video generation \citep{latentDiff,Imagen,classifierfree}. Generally, these methods typically involve learning continuous transformation trajectories that map a simple initial distribution (i.e. Gaussian noise) to the target data distribution (i.e. images) by solving ODEs (Figure \ref{fig1}).

Among those ODE-based models, the diffusion models have attracted the most attention due to their exceptional ability to generate realistic samples. The diffusion models add noise through a forward diffusion process and gradually remove it in a reverse process that can be reformulated as a probability flow ODE (PF-ODE)\citep{score}. Despite their success, PF-ODE-based diffusion models face drawbacks due to their iterative nature, leading to high computational costs and prolonged sampling times during inference.

Another ODE-based approach, rectified flow models \citep{rectified, lipman2022flow}, focused on learning smoother ODE trajectories to reduce truncation errors during numerical integration. By reducing the curvature of the generative paths, rectified flow enhances sampling efficiency and decreases the computational burden. However, even with smoother trajectories, rectified flow models still require considerable iterations or reflow to produce high-quality samples. 

Inspired by concepts from electrostatics, Poisson Flow Generative Models (PFGM) ++ \citep{pfgm,xu2023pfgm++}, developed the flow-based generative process that solves an ODE derived from the Poisson equation, tracing a path from a simple initial distribution (e.g., noise on a large hemisphere) to the target data distribution residing on a lower dimensional hyperplane. Like the diffusion models and rectified flow, PFGM++ requires multiple steps during inference.

All aforementioned methods required multiple steps of solving different ODE functions in inference to obtain high-quality results. Furthermore, the ODE-based models naturally inherit the discretization error of numerical solvers of ODE, which restricts the quality of samples when only a few numbers of function evaluations (NFEs) are used or struggle with training instability when neural ODEs are used to approximate the ODE solution using neural networks. Given these challenges associated with ODE-based models, a natural question arises: can we learn the result of ODE-based trajectory paths directly without solving the ODE functions? Therefore, we can take the ODE function-defined generative model and solve this without an ODE solver. The answer is yes. Here we proposed \textbf{Integration Flow}, to the best of our knowledge, the first model with the unified structure to estimate ODE-based generative models.

Integration Flows represent a new type of generative model. Unlike traditional ODE-based approaches that focus on approximating the instantaneous drift term of an ODE or depend on iterative sampling methods, Integration Flows directly estimate the integrated effect of the cumulative transformation dynamics over time. This holistic approach allows for the modeling of the entire generative path in a single step, bypassing the accumulation of errors associated with high-curvature trajectories and multiple function evaluations. Integration Flows do not employ the ODE solver and eliminate the need for multiple sampling iterations, significantly reducing computational costs and enhancing efficiency. 

Furthermore, to increase the training stability and accuracy in reconstructing, Integration Flow explicitly incorporates the target state $\mathbf{x}_0$ as the anchor state in guiding the reverse-time dynamics from the intermediate state $\mathbf{x}_t$. We have theoretically proven that incorporating the target state $\mathbf{x}_0$ as the anchor state can provide a better or at least equal accurate estimation of $\mathbf{x}_0$.

In summary, Integration Flows addresses the limitations of existing ODE-based generative models by providing a unified and efficient approach to model the transformation between distributions. Our contributions can be outlined as follows:
\begin{itemize}
\item Introduction of Integration Flows: We present Integration Flows, a novel generative modeling framework that estimates the integrated dynamics of continuous-time processes without relying on iterative sampling procedures or traditional ODE solvers, that is, it supports one-step generation for (any) well-defined ODE-based generative models.
\item Unified ODE-Based Generative Modeling: Integration Flows can adapt different ODE-based generative processes, offering flexibility and unification across different generative modeling approaches.
\item Enhanced Sampling Efficiency and Scalability: Through empirical evaluations, we show that Integration Flows achieve improved sampling efficiency and scalability compared to existing ODE-based models, such as diffusion models, Rectified Flows, and PFGM++. In particular, we set the state-of-the-art performance for one-step generation using Rectified Flow and PFGM++.  
\end{itemize}

 \section{Related Works}

Our approach is related to existing works aiming to address the drawbacks of iterative inference in traditional diffusion models\cite{ddpm,score}. Consistency Models (CMs)\cite{consistency}, for instance, learn a direct mapping from the noisy data distribution to clean samples in a way that is consistent across different noise levels. 
One major drawback of CM is that it requires enhanced training techniques to achieve good results\cite{iCT}. These improved techniques are specifically tailored to the VE diffusion model, lacking generalizability to other generative models such as Rectified Flow or those employing different noise schedulers like Variance Preserving (VP). This limits the flexibility and extensibility of CMs in broader applications. 

Adapting CM to other generative models needs a special redesign of the model. For example, Consistency-FM\cite{yang2024consistency} adapted the concept of CM and directly defines straight flows that start from different times to the same endpoint.  Consistency-FM improved the quality of samples moderately to the original FM, which is still far below ours. Moreover, 2-Rectified Flow++\cite{lee2024improving} uses a similar framework of CMs and incorporates different loss and time scheduling and obtain an improved performance on 2-rectified flow. However, reflow needs more computational resources. 

Consistency Trajectory Models (CTMs)\cite{ctm} introduce adversarial training to improve sample fidelity. However, adversarial training is notoriously difficult to train, often leading to issues such as mode collapse and training divergence\cite{sauer2023stylegantunlockingpowergans}. As a result, CTM reduces training stability. 

Directly Denoising Diffusion Models (DDDMs)\cite{zhang2024directly}, are the most relevant to our work but are limited to the VP case and can be seen as a special case of Integration Flow Models. Unlike DDDMs, our method provides a generalized approach that can be seamlessly extended to various generative models and noise schedulers without the need for model-specific training techniques. Integration Flow offers a more versatile and efficient solution for high-quality sample generation in a single step.


\section{Method}
\label{sec:method}
In this section, we will introduce the  Integration Flow Models based on the general form of ODE-based generative models.
\subsection{general form of ODE-based generative models}
Consider an initial state $\mathbf{x}_T$ drawn from a distribution $p(\mathbf{x}_T)$, typically chosen to be a simple distribution such as a Gaussian.
The goal is to estimate $\mathbf{x}_0$, which is aligned with the data distribution $p_{\text{data}}$, by mapping $\mathbf{x}_T$ back through a continuous transformation process. Let $\{\mathbf{x}_s\}_{s=0}^{T}$ represent a continuous transformation trajectory from $\mathbf{x}_T$ to $\mathbf{x}_0$, where $\mathbf{x}_s$ denotes the state at intermediate time $s \in [0, T]$. To describe the reverse-time dynamics that map $\mathbf{x}_T$ back to $\mathbf{x}_0$, we define a reverse-time ODE:
\begin{equation} \frac{d\mathbf{x}_s}{ds} = v(\mathbf{x}_s, s), \label{eq 1
} \end{equation}
where $v: \mathbb{R}^n \times [0, T] \rightarrow \mathbb{R}^n$ is a continuous function defining the system's dynamics in reverse time.

The process of obtaining $\mathbf{x}_0$ from $\mathbf{x}_T$ involves solving this reverse-time ODE, which can be understood as computing the integral: 
\begin{equation}
\begin{split}
\int_T^0 \frac{\mathrm{~d} \mathbf{x}_s}{\mathrm{~d} s} \mathrm{~d} s=\int_T^0 v(\mathbf{x}_s, s)\mathrm{~d} s \Longleftrightarrow \\ \mathbf{x}_0=\mathbf{x}_T+\int_T^0 v(\mathbf{x}_s, s) \mathrm{~d} s \label{eq 2} 
\end{split}
\end{equation}
The solution of the reverse-time ODE aligns marginally in distribution with the forward process, meaning that the distribution of $\mathbf{x}_0$ obtained by solving the reverse ODE starting from $\mathbf{x}_T \sim p_T(\mathbf{x})$ approximates the target distribution $p_{data}$.

While traditional ODE solvers and neural ODE methods are commonly used to solve the \eqref{eq 2}, they have notable drawbacks. The numerical solvers of ODE can not avoid the discretization error \citep{bridge}, which restricts the quality of samples when only a few NFEs are used. Second, neural ODEs \citep{chen2018neural}, which approximate the ODE solution using neural networks, face a high challenge during gradient backpropagation due to their high memory consumption \citep{gholami2019anode}.

\subsection{Integration Flows}

To overcome the challenges associated with ODE solvers and neural ODEs, we propose Integration Flow to directly estimate the integrated effect of continuous-time dynamics. The cumulative effect of the reverse-time dynamics over the interval $[0, t]$, which is defined in integral $\int_t^0 v(\mathbf{x}_s, s) ds $, can be obtained as:
\begin{equation}\int_t^0 v(\mathbf{x}_s, s) ds=V\left(\mathbf{x}_0, 0\right)-V\left(\mathbf{x}_t, t\right)  \end{equation}
where $V(\mathbf{x}_s, s)$ is an antiderivative of $v(\mathbf{x}_s, s)$ with respect to $s$.
Then, we defined a function $G(\mathbf{x}_0, \mathbf{x}_t, t)$ as follows:
\begin{equation}G(\mathbf{x}_0, \mathbf{x}_t, t) := V\left(\mathbf{x}_t, t\right)-V\left(\mathbf{x}_0, 0\right)\end{equation}
This function encapsulates the total influence of the dynamics from an intermediate time $t$ to the final time $0$, which leads to the equation:
\begin{equation} 
\mathbf{x}_0  = \mathbf{x}_t-G(\mathbf{x}_0, \mathbf{x}_t, t).\end{equation}
Next, we define the function: 
\begin{equation} g(\mathbf{x}_0, \mathbf{x}_t, t) := \mathbf{x}_t - G(\mathbf{x}_0, \mathbf{x}_t, t). \label{eq 3} \end{equation}
Therefore, we have: 
\begin{equation} \mathbf{x}_0 = g(\mathbf{x}_0, \mathbf{x}_t, t). \label{eq 4} \end{equation} 
Thus, $g(\mathbf{x}_0, \mathbf{x}_t, t)$ is the \textbf{solution} of the reversed time ODE from initial time $t$ to final time $0$, which encapsulates the cumulative effect of the reverse dynamics from the initial time $t$ to the final time $0$, providing an accumulation description of how the target state $\mathbf{x}_0$ transformed from the intermediate state $\mathbf{x}_t$. 

Integration Flow explicitly incorporates the target state $\mathbf{x}_0$ as the anchor state in guiding the reverse-time dynamics from the intermediate state $\mathbf{x}_t$, which contributes to both stability and accuracy in reconstructing $\mathbf{x}_0$ from intermediate states $\mathbf{x}_t$. Since Integration Flow bypasses the ODE solver, it provides a unified framework for ODE-based generative models, allowing for one-step generation across a variety of processes.

 \subsection{Neural Network Approximation}
We can implement $G$ using a neural network parameterized by $\boldsymbol{\theta}$ \cite{zhang2024directly}. The approximated predictive model is thus defined as:
\begin{equation}g_{\boldsymbol{\theta}}\left(\mathbf{x}_0, \mathbf{x}_t, t\right)=\mathbf{x}_t-G_{\boldsymbol{\theta}}\left(\mathbf{x}_0, \mathbf{x}_t, t\right)\label{eq 5}\end{equation}
where $G_{\boldsymbol{\theta}}$ approximates $G$. 

Equation 8 works well when noise variance is scaled to unit variance, namely Variance Preserving (VP) of diffusion model \cite{zhang2024directly}. However, if the noise variance is large, such as the Variance Exploring (VE) cases of diffusion model or the PFGM++ model (shown in section 4), the neural network becomes unstable because it needs to adjust its output to cancel the large noise. To ensure the stability of the integration flow, we reparametrized the dynamics $g_{\boldsymbol{\theta}}(\mathbf{x}_0, \mathbf{x}_t, t)$ as the following:
\begin{equation} g_{\boldsymbol{\theta}}(\mathbf{x}_0, \mathbf{x}_t, t) = a_t \mathbf{x}_t + b_t G_{\boldsymbol{\theta}}(\mathbf{x}_0, \mathbf{x}_t, t). \label{eq 6} \end{equation}
where $a_t$ and $b_t$ are time-dependent scalar functions designed to modulate the contributions of $\mathbf{x}_t$ and $G(\mathbf{x}_0, \mathbf{x}_t, t)$, respectively. This formulation introduces better flexibility in the evolution of the integration flow over time, particularly in scenarios where the straightforward application (Equation \eqref{eq 5}) of integration may introduce instability, especially when the magnitudes of the intermediate state $\mathbf{x}_t$ become large. 

The goal of the training neural network is to accurately recover $\mathbf{x}_0$ from $\mathbf{x}_t$. Therefore, it is essential that:
\begin{equation}
g_{\boldsymbol{\theta}}\left(\mathbf{x}_0, \mathbf{x}_t, t\right) \approx g\left(\mathbf{x}_0, \mathbf{x}_t, t\right) = \mathbf{x}_0
\end{equation}
However, $g_{\boldsymbol{\theta}}\left(\mathbf{x}_0, \mathbf{x}_t, t\right)$ may not directly recover the true $\mathbf{x}_0$ accurately. We then employed an iterative refinement process. Starting with an initial $\mathbf{x}_0^{(0)} \sim \mathcal{N}(\mathbf{0}, \mathbf{I})$, the estimate is progressively refined using the following update rule:
\begin{equation}
\mathbf{x}_0^{(n+1)} = g_{\boldsymbol{\theta}}\left(\mathbf{x}_0^{(n)}, \mathbf{x}_t, t\right)=a_t\mathbf{x}_t+ b_tG_{\boldsymbol{\theta}}\left(\mathbf{x}_0^{(n)}, \mathbf{x}_t, t\right)
\end{equation}
Through this iterative process, the neural network will effectively minimize the discrepancy between the iteratively estimated $\mathbf{x}_0^{(n)}$ and the true initial state $\mathbf{x}_0$ with properly defined loss.

\subsection{Theoretical justification}

\textbf{Theorem 1 (Stability)}: Let $\mathbf{x}_0$ represent the target state, $\mathbf{x}_t$ represent an intermediate state, and $t$ represent the time. Let $\mathbf{x}_0^{(n)}$ be an auxiliary estimate of $\mathbf{x}_0$ obtained through an iterative process. Consider the following two estimators:

(a)$\hat{\mathbf{x}}_0=g^{\prime}_{\boldsymbol{\theta}}\left(\mathbf{x}_t, t\right)$, which estimates $\mathbf{x}_0$ based only on $\mathbf{x}_t$ and $t$, analogous to $\mathbb{E}\left[\mathbf{x}_0 |\mathbf{x}_t\right]$;

(b) $\tilde{\mathbf{x}}_0 = g_{\boldsymbol{\theta}}\left(\mathbf{x}_0^{(n)}, \mathbf{x}_t, t\right)$, which estimates $\mathbf{x}_0$ based on both $\mathbf{x}_0^{(n)}, \mathbf{x}_t$, and $t$, analogous to $\mathbb{E}[\mathbf{x}_0|\mathbf{x}_t, \mathbf{x}_0^{(n)}]$.

Then, the estimator $\tilde{\mathbf{x}}_0$, which includes additional conditional information $\mathbf{x}_0^{(n)}$, provides a more accurate estimation of $\mathbf{x}_0$ compared to $\hat{\mathbf{x}}_0$, in terms of mean squared error (MSE). That is,
\begin{equation}
\mathbb{E}\left[\left\|\mathbf{x}_0-\tilde{\mathbf{x}}_0\right\|^2\right] \leq \mathbb{E}\left[\left\|\mathbf{x}_0-\hat{\mathbf{x}}_0\right\|^2\right]
\end{equation}

Theorem 1 justifies that the estimator $\tilde{\mathbf{x}}_0$ is at least as accurate as $\hat{\mathbf{x}}_0$ under the same convex metric $d(\cdot, \cdot)$, illustrating that $\tilde{\mathbf{x}}_0 = g_{\boldsymbol{\theta}}\left(\mathbf{x}_0^{(n)}, \mathbf{x}_t, t\right)$ provides a better or at least equal estimate compared to $\hat{\mathbf{x}}_0=g^{\prime}_{\boldsymbol{\theta}}\left(\mathbf{x}_t, t\right).$

\textbf{Theorem 2 (Non-Intersection)}: Suppose that the neural network is sufficiently trained and $\boldsymbol{\theta}^{*}$ is obtained, such that: $g_{\boldsymbol{\theta}^*}(\mathbf{x}_0^{(n)},\mathbf{x}_t, t) \equiv g(\mathbf{x}_0,\mathbf{x}_t, t)$ for any $t \in [0, T]$ and $\mathbf{x}_0$ sampled from $p_{\text{data}}$, and
$v\left(\mathbf{x}_s, s\right)$ meets Lipschitz condition. 

Then for any \(t \in[0, T]\), the mapping \(g_{\boldsymbol{\theta}^*}(\mathbf{x}_0^{(n)},\mathbf{x}_t, t): \mathbb{R}^N \rightarrow \mathbb{R}^N\) is bi-Lipschitz. Namely, for any \(\mathbf{x}_t, \mathbf{y}_t \in \mathbb{R}^N\)
\begin{equation}
\begin{split}
e^{-Lt}\left\|\mathbf{x}_t-\mathbf{y}_t\right\|_2 &\leq \left\|g_{\boldsymbol{\theta}^*}
(\mathbf{x}_0^{(n)},\mathbf{x}_t, t) - g_{\boldsymbol{\theta}^*}(\mathbf{y}_0^{(n)},\mathbf{y}_t, t)\right\|_2 \\
&\leq e^{Lt}\left\|\mathbf{x}_t-\mathbf{y}_t\right\|_2 .
\end{split}
\end{equation}
This implies that if given two different starting points, say \(\mathbf{x}_T\neq \mathbf{y}_T\), by the bi-Lipschitz above, it can be concluded that \(g_{\boldsymbol{\theta}^*}(\mathbf{x}_0^{(n)},\mathbf{x}_T, T)\neq g_{\boldsymbol{\theta}^*}(\mathbf{y}_0^{(n)},\mathbf{y}_T, T)\) i.e., \(\mathbf{x}_0^{(n+1)} \neq \mathbf{y}_0^{(n+1)}\), which indicates the reverse path of Integration Flow does not intersect.

\textbf{Theorem 3 (Integration Flow Optimality)}: Integration Flow is optimal for Flow Matching/Rectified Flow in the sense that any solution $v$ to the continuous-time problem has a cost no better than the cost of the integrated approach. Therefore, a direct one-step method (via $G$) achieves the best possible value.

The proof of Theorems is presented in the Appendix\ref{AppB}.

\section{Integration Flow for different ODE-based generative models}
\label{sec:app}
\begin{table*}[t!]
\setlength\tabcolsep{2pt}  
\caption{The different design choice of Integration Flow for different ODE-based methods. For training, we use the discrete time steps with $T=1000.$ }
\label{tab:comparison}
\centering
\small

\begin{tabular}{p{4cm}p{4cm}p{4.5cm}p{4cm}}
\hline
&VE\citep{score} & Rectified Flow\citep{rectified}& PFGM++\cite{xu2023pfgm++}  \\
\hline
\multicolumn{4}{l}{\textbf{Training }} \\
Noise scheduler & $\sigma_{\min }\left(\frac{\sigma_{\max }}{\sigma_{\min }}\right)^{t/T}$ & Linear interpolation, $\sigma=0$ &$R_t \mathbf{v}_t$, where $\sigma_t\sim p(\sigma_t)$,\\& & & $r_t=\sigma_t \sqrt{D}$, $R_t \sim p_{r_t}(R),$ \\
& & & $\mathbf{v}_t=\frac{\mathbf{u}_t}{\left\|\mathbf{u}_t\right\|_2}$,$\mathbf{u}_t \sim$ $\mathcal{N}(\mathbf{0}, \mathbf{I})$ \\

 Steps & $ t\in [1,2,...,T]$  & $t\sim \text{Uniform}[0,1]$ & $ t\in [1,2,...,T]$ \\
\hline
\multicolumn{4}{l}{\textbf{Network and preconditioning}} \\
Architecture of $F_\theta$ & ADM &  ADM &  ADM  \\
$a_t$ & $\sigma_{\min}/\sigma_t $ & $0$ &  $\sigma_{\min}/R_t $ \\
$b_t$ & $1-\sigma_{\min}/\sigma_t$ & $1$ & $1-\sigma_{\min}/R_t$ \\

\hline
\multicolumn{4}{l}{\textbf{Sampling}} \\ One step & &$\mathbf{x}_0^{(\text{est})}=a_t \mathbf{x}_t+b_t G_{\boldsymbol{\theta}}\left(\mathbf{x}_0^{(0)}, \mathbf{x}_t, t\right)$\\
Multistep $n$  & &$\mathbf{x}_0^{(\text{est})}=a_t \mathbf{x}_t+b_t G_{\boldsymbol{\theta}}\left(\mathbf{x}_0^{(n)}, \mathbf{x}_t, t\right)$ &  \\
\hline
\multicolumn{4}{l}{\textbf{Parameters}} \\
& $\sigma_{\text{min}} = 0.01$ & & $\sigma_{\text{min}} = 0.01$  \\
& $\sigma_{\text{max}} = 50$ & ------ & $\sigma_{\text{max}} = 50$  \\
 & & &$D =2048$\\

\hline
\end{tabular}%

\end{table*}
In this section, we apply Integration Flow to three ODE-based generative models. More training settings can be seen in Table \ref{tab:comparison}.

\textbf{Integration Flow for Variance Exploding (VE) Diffusion Model}.
The forward process in the Variance Exploding (VE) diffusion model\citep{score,edm} adds noise to the data progressively. This process is described as:
\begin{equation}
\mathbf{x}_t=\mathbf{x}_0+\sigma_t \boldsymbol{\epsilon}, \quad t \in[0, T]
\end{equation}
where $\mathbf{x}_0 \sim p_{\text {data }}$,$\sigma_t$ denotes noise schedule that increases with time $t$, $\boldsymbol{\epsilon} \sim \mathcal{N}(0, \mathbf{I}).$ 
The reverse process aims to denoise the data by starting from a noisy sample $\mathbf{x}_T$ and evolving it back to the clean data distribution $p_{\text {data }}$. This is achieved using the PF-ODEs, which model the continuous denoising process in the reverse direction. The PF-ODE is given by:
\begin{equation}
\frac{\mathrm{d} \mathbf{x}_t}{\mathrm{~d} t}=-\frac{1}{2} \frac{\mathrm{~d} \sigma_t^2}{\mathrm{~d} t} \nabla_{\mathbf{x}_t} \log p_t\left(\mathbf{x}_t\right)
\end{equation}
where $\nabla_{\mathbf{x}_t} \log p_t\left(\mathbf{x}_t\right)$ is the score function, representing the gradient of the log-probability of the data distribution $p_t\left(\mathbf{x}_t\right)$ at time $t$. 
We adopt the noise scheduler as $\sigma_{\min }\left(\frac{\sigma_{\max }}{\sigma_{\min }}\right)^{t/T}$, where noise increases exponentially over time from $\sigma_{\min }$ to $\sigma_{\max }$, and time step $t$ is designed as $ t\in [1,2,...,T]$. Therefore, the integration flow can be expressed as: 
\begin{equation}
 g_{\boldsymbol{\theta}}\left(\mathbf{x}_0, \mathbf{x}_t, t\right)=\frac{\sigma_{\min }}{\sigma_t}\mathbf{x}_t+\left(1-\frac{\sigma_{\min }}{\sigma_t}\right) G_{\boldsymbol{\theta}}(\mathbf{x}_0, \mathbf{x}_t, t)
\end{equation}
where the preconditioning terms are set as $a_t = \sigma_{\min } / \sigma_t$, and $b_t = 1-\sigma_{\min } / \sigma_t$, which modulate the network's response to different noise levels throughout training. The detailed derivation of the Integration Flow for VE diffusion model is in Appendix\ref{App1}.

\textbf{Integration Flow for Rectified Flows}. Rectified flows \citep{rectified,albergo2022building,lipman2022flow} uses linear interpolation to connect the data distribution $p_{data}$ and a standard normal distribution $p_{\mathbf{z}}$ by introducing a continuous forward process that smoothly transitions between these two distributions, which is defined as:
\begin{equation}
\mathbf{z}_t=(1-t) \mathbf{x}_0+t \mathbf{z}
, \quad t \in [0,1] \end{equation}
where $\mathbf{x}_0$ is a sample drawn from the data distribution $p_{data}, \mathbf{z} $ is sampled from the standard normal distribution. Time step $t$ is sampled from $\text{Uniform}[0,1].$ Since this is a deterministic linear interpolation, there is no need of a noise scheduler. This interpolation ensures that at $t=0$, it recovers the original data point, i.e., $\mathbf{z}_0=\mathbf{x}_0$, and at $t=1$, the point has been mapped entirely into the noise distribution, i.e., $\mathbf{z}_1=\mathbf{z}$. Thus, a straight path is created between the data and the noise distributions. 

Liu et al.\citep{rectified} demonstrated that for $\mathbf{z}_0 \sim p_{\mathbf{x}}$, the dynamics of the following ODE produce marginals that match the distribution of $\mathbf{x}_t$ for any $t$ :
\begin{equation}
\frac{d \mathbf{z}_t}{d t}=v\left(\mathbf{z}_t, t\right)
\end{equation}
Since the interpolation ensures that $\mathbf{x}_1=\mathbf{z}$, the forward ODE transports samples from the data distribution $p_{\mathbf{x}}$ to the noise distribution $p_{\mathbf{z}}$. To reverse this process, starting with $\mathbf{z}_1 \sim p_{\mathbf{z}}$, the ODE can be integrated backward from $t=1$ to $t=0$, ultimately reconstructing samples from the data distribution.

The integration flow of Rectified Flows can be expressed as:
\begin{equation}
\mathbf{x}_0=g_{\boldsymbol{\theta}}\left(\mathbf{x}_0, \mathbf{z}_t, t\right) =G_{\boldsymbol{\theta}}\left(\mathbf{x}_0, \mathbf{z}_t, t\right)
\end{equation}
Equivalent to $a_t =0, b_t = 1$ in \eqref{eq 6}. The detailed derivation of Integration Flow for Rectified Flow is in the Appendix\ref{App2}. Moreover, Integration Flow supports Stochastic Interpolants \cite{albergo2022building} as well.

\begin{figure*}
    \centering
    \includegraphics[scale=0.7]{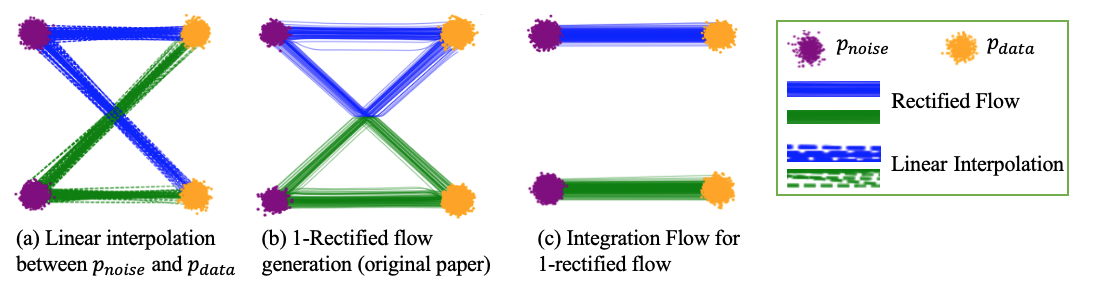}
    \caption{Rectified flow process. (a)-(b) Flow rewires trajectories of original rectified flow, regenerated from \cite{rectified}. (c) Straighten flows of Integration Flow based 1-rectified flow. }
    \label{fig:rect}
\end{figure*}

\textbf{Integration Flow PFGM++}. PFGM++ \citep{xu2023pfgm++} is a generalization of PFGM\citep{pfgm} that embeds generative paths in a high-dimensional space. It reduces to PFGM when $D=1$ and to diffusion models when $D \rightarrow \infty$.

In PFGM++, each data point $\mathbf{x} \in \mathbb{R}^N$ is augmented by additional variables $\mathbf{z}=\left(z_1, \ldots, z_D\right) \in$ $\mathbb{R}^D,$ resulting in an augmented data representation $\tilde{\mathbf{x}}=(\mathbf{x}, \mathbf{z}) \in \mathbb{R}^{N+D}$. Due to the rotational symmetry of the electric field in the augmented space, the problem can be simplified by considering only the radial norm $r=\|\mathbf{z}\|_2$. This reduces the augmented data representation to $\tilde{\mathbf{x}}=(\mathbf{x}, r)$, where $r$ acts as a scalar anchor variable.

PFGM++ uses the electric field $\mathbf{E}(\tilde{\mathbf{x}})$ to drive the dynamics of the generative process. 
Using the radial symmetry of the electric field, the backward ODE that governs the generative process can be expressed as:
\begin{equation}    
\frac{d \mathbf{x}}{d r}=\frac{\mathbf{E}(\tilde{\mathbf{x}})_{\mathbf{x}}}{E(\tilde{\mathbf{x}})_r}
\end{equation}
By solving this ODE in reverse, one can transport points from the high dimensional augmented space back to the original data space, completing the generative process.

PFGM++ introduces an alignment method to transfer hyperparameters from diffusion models (where $D \rightarrow \infty)$ to finite-dimensional settings. The alignment is based on the relationship:
\begin{equation}
r=\sigma \sqrt{D}
\end{equation}
This formula ensures that the phases of the intermediate distributions in PFGM++ are aligned with those of diffusion models. The relation allows transferring finely-tuned hyperparameters like $\sigma_{\max }$ and $p(\sigma)$ from diffusion models to PFGM++ using:
\begin{equation}
r_{\max }=\sigma_{\max } \sqrt{D}, \quad p(r)=\frac{p(\sigma=r / \sqrt{D})}{\sqrt{D}}
\end{equation}
Further, \citep{xu2023pfgm++} showed \begin{equation}\frac{d \mathbf{x}}{d r}=\frac{d \mathbf{x}}{d \sigma}\end{equation}
where $\sigma$ changes with time. Thus, we adopt the noise scheduler the same as in VE case. And the perturbation to the original data $\mathbf{x}_0$ can be written as: 
\begin{equation}
\mathbf{x}_t = \mathbf{x}_0+R_t \mathbf{v}_t
\end{equation}
Specifically, for each data point $\mathbf{x}_0$, a radius $R_t$ is sampled from the distribution $p_{r_t}(R)$(See Appendix B in \citep{xu2023pfgm++} to sample $R_t$). To introduce random perturbations, uniform angles are sampled by first drawing from a standard multivariate Gaussian, $\mathbf{u}_t \sim \mathcal{N}(\mathbf{0}, \mathbf{I})$, and then normalizing these vectors to obtain unit direction vectors $\mathbf{v}_t=\frac{\mathbf{u}_t}{\left\|\mathbf{u}_t\right\|_2}$. This perturbation acts as a forward process in PFGM++, analogous to the forward process in diffusion models. 

The Integration Flow $g_{\boldsymbol{\theta}}\left(\mathbf{x}_0, \mathbf{x}_t, \sigma_t\right)$ of PFGM++ can be expressed as:
\begin{equation}
g_{\boldsymbol{\theta}}(\mathbf{x}_0, \mathbf{x}_t, t) = \frac{\sigma_{\min }}{R_t} \mathbf{x}_t + (1-\frac{\sigma_{\min }}{R_t} )G_{\boldsymbol{\theta}}(\mathbf{x}_0, \mathbf{x}_t, t)  \end{equation} 
with $a_t = \sigma_{\min }/R_t$ and $b_t = 1-\sigma_{\min }/R_t,$ and the detailed derivation of $a_t, b_t$ is shown in Appendix\ref{App3}. 
\section{Experiments}
\label{sec:exp}

\begin{figure*}[t!]
    \centering
    \includegraphics[scale=0.59]{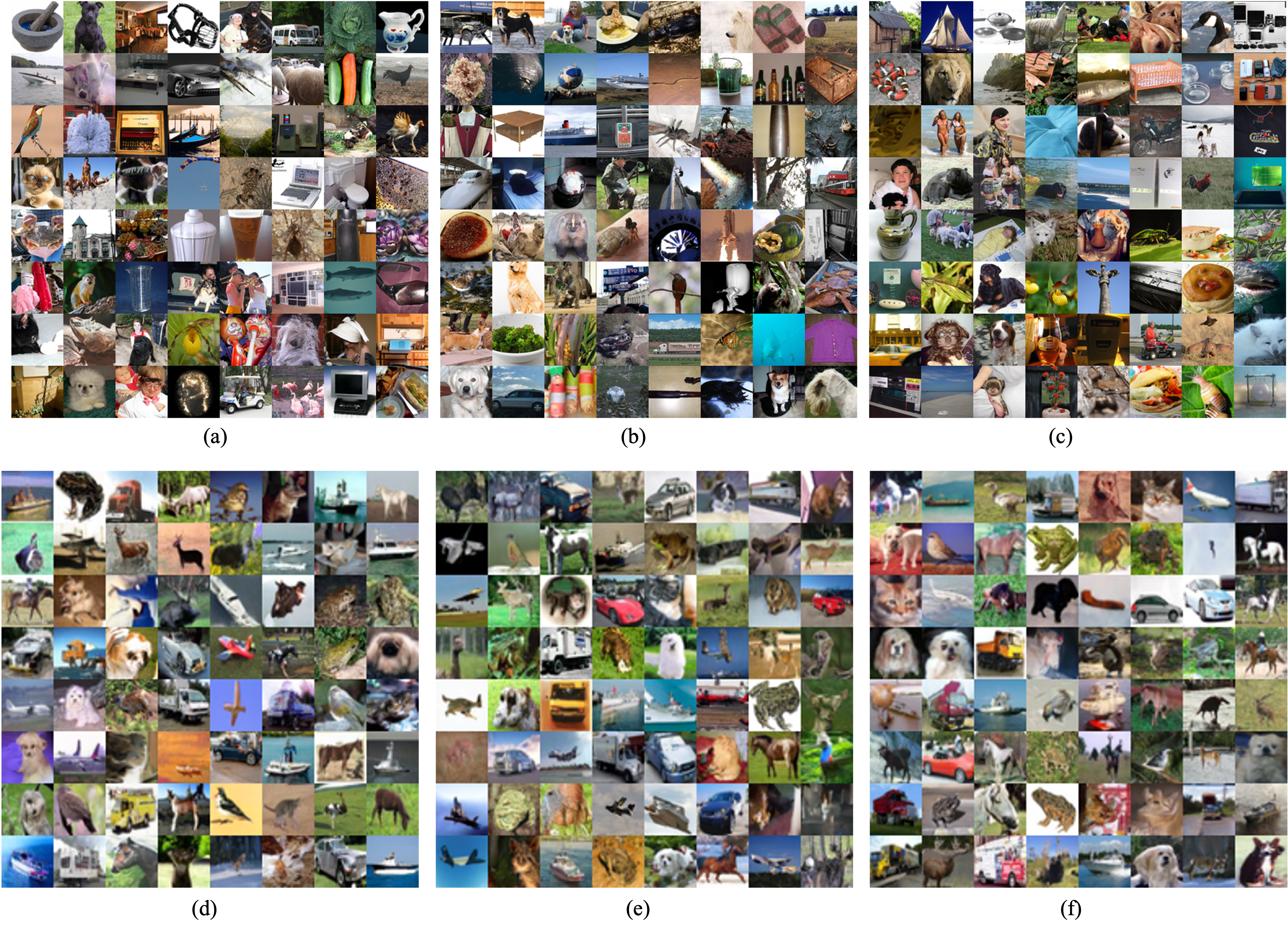}
    \caption{One-step samples from Integration Flow. Top: ImageNet; Bottom: CIFAR-10. (a) and (d): 1-rectified flow with FID 4.35 and 3.36; (b) and (e): VE with FID 4.09 and 2.86; (c) and (f): PFGM++ with FID 4.15 and 2.91}
    \label{fig:combined}
\end{figure*}

To evaluate our method for image generation, we train several Integration Flow Models on CIFAR-10 \cite{CIFAR} and ImageNet 64x64 \cite{imagenet}
and benchmark their performance with
competing methods in the literature. Results are
compared according to Frechet Inception Distance (FID, \citet{FID}), which is computed between 50K generated samples and the whole training set. The training and sampling algorithm can be found in the Appendix\ref{App}.

\subsection{Implementation Details}
\textbf{Architecture.} We employed the ADM architecture \cite{ADM} for the implementation of Integration Flow. For CIFAR-10, the base channel dimension is set to 128 and is multiplied by 1,2,2,2 in four stages. For ImageNet 64x64, the base channel dimension is set to 192 and is multiplied by 1,2,3,4 in four stages. For both datasets, we used three residual blocks per stage and set the Dropout \cite{dropout} rate to 0.3. We also used cross-attention modules at both the 16x16 resolution and the 8x8 resolution, where the conditioning image $\mathbf{x}_0^{(n)}$ is integrated into the network. The models for CIFAR-10 are unconditional, but the models for ImageNet are conditioned on class labels. 

\textbf{Loss function} Inspired by \cite{iCT}, we adopt the Pseudo-Huber metric family \cite{PH} as the loss function, defined as
\begin{equation}
    d(\boldsymbol{x},\boldsymbol{y}) = \sqrt{\|\boldsymbol{x}-\boldsymbol{y}\|_2^2+c^2}-c
\end{equation}
where $c$ is an adjustable hyperparameter. The Pseudo-Huber metric is more robust to outliers compared to the squared $\ell_2$ loss metric because it imposes a smaller penalty for large errors, while still behaving similarly to the squared $\ell_2$ loss metric for smaller errors. We set  $c=0.00016$ for VE, $c=0.0001$ for Rectified Flow, and $c=0.00016$ for PFGM++, respectively.

\textbf{Other settings.} We used Adam for all of our experiments. For VE and PFGM++, we set T=1000 for CIFAR10 and T=2000 for ImageNet. For all three Integration Flow models, we trained 400,000 iterations with a constant learning rate of 0.0002 and batch size of 128 for CIFAR 10 and trained 600,000 iterations with a constant learning rate of 0.0001 and batch size of 1024 for ImageNet. All experiments used the exponential moving average (EMA) of the weights during training with a decay factor of 0.9999. All models are trained on NVIDIA H100 DGX.  


\subsection{Comparison to SOTA} 

We compare our models against the state-of-the-art generative models on CIFAR-10 and ImageNet 64X64. The quantitative results are summarized in Table \ref{tab:cifar} and Table \ref{tab:imagenet}. For VE, Integration Flow achieved performances that are comparable to the state-of-the-art on both datasets. Specifically, Integration Flow one-step generation obtains FIDs of 2.86 for CIFAR-10 and 4.09 for ImageNet.  

For Rectified Flow, the one-step generation with Integration Flow has reached FIDs of 3.36 for CIFAR10 and 4.35 for ImageNet without reflow, which is the state-of-the-art performance in the Rectified Flow. Generally, the Rectified Flow needs to be applied at least twice (Reflow) to obtain a reasonable one-step generation performance \citep{rectified,liu2023instaflow}. The Integration Flow one-step generation has a better (CIFAR10) or comparable (ImageNet) performance compared to the best results of Reflow. As shown in Figure \ref{fig:rect}, the flow has already become exactly straight for Integration Flow-based 1-rectified flow. This explains why Integration Flow-based 1-rectified flow can achieve good results as Reflow did. 

For PFGM++, Integration Flow one-step generation has reached FIDs of 2.91 for CIFAR10 and 4.15 for ImageNet. We are the first to show that the PFGM++ can also achieve one-step generation with good performance.

\begin{table}[t!]
\setlength\tabcolsep{0.01cm}
    \centering
    \caption{Performance evaluation of unconditional samples on CIFAR-10}
    \fontsize{8.5pt}{8.5pt}\selectfont
    \begin{tabular}{lccc}
    \toprule
         Method & NFE($\downarrow$) & FID($\downarrow$) & IS($\uparrow$) \\
         \toprule
         \multicolumn{4}{l}{\textbf{Diffusion Models - Fast Samplers \& Distillation}} \\
         DDIM \cite{ddim}& 10 & 13.36 & \\
         DPM-solver-fast \cite{dpm-solver}& 10 & 4.70 \\
         3-DEIS \cite{3deis}& 10 & 4.17 \\
         UniPC \cite{unipc}& 10 & 3.87 \\
         DFNO (LPIPS) \cite{DFNO} &1 &3.78 \\
         Knowledge Distillation \cite{knowledge} & 1 & 9.36\\
         TRACT \cite{tract}&1 &3.78 \\
         &2 &3.32 \\
         Diff-Instruct \cite{instruct}&1 &4.53 &9.89 \\
         \toprule
          \multicolumn{4}{l}{\textbf{Diffusion Models - Direct Generation}}\\
         
         Score SDE \cite{score}&2000 &2.38 &9.83 \\
         Score SDE (deep) \cite{score}  &2000 &2.20 &9.89 \\
DDPM \cite{ddpm} &1000 &3.17 &9.46 \\
LSGM  \cite{LSGM}&147 &2.10 \\
EDM   \cite{edm}&35  &2.04 &9.84 \\
EDM-G++ \cite{edmG}&35 &1.77 \\
 \toprule
 \multicolumn{4}{l}{\textbf{GAN Models}}\\
BigGAN  \cite{bigGAN}&1 &14.7 &9.22 \\
StyleGAN2 \cite{style2}&1 &8.32 &9.21 \\
StyleGAN2-ADA \cite{stylegan2-ada}  &1 &2.92 &9.83 \\
\toprule
\multicolumn{4}{l}{\textbf{Consistency Models}}\\
         
 CD (LPIPS) \cite{consistency} &1 &3.55 &9.48 \\
         &2 &2.93 &9.75 \\
CT (LPIPS) \cite{consistency}&1 &8.70 &8.49 \\
&2 &5.83 &8.85 \\
iCT \cite{iCT} &1 &2.83 &9.54 \\
&2 &2.46 &9.80 \\
iCT-deep \cite{iCT} &1 & 2.51 & 9.76 \\
&2 &2.24 &9.89 \\
CTM \cite{ctm} & 1 &5.19 \\
CTM \cite{ctm} + GAN &1 &2.39 \\
\toprule
\multicolumn{4}{l}{\textbf{Rectified Flows}}\\
1-Rectified flow \cite{rectified} &1 &378 &1.03\\
1-Rectified flow (+distill) \cite{rectified} &1 &6.18 &9.08\\
2-Rectified Flow (+distill)\cite{rectified} &1 &4.85 &9.01\\
2-Rectified flow++ \cite{lee2024improving} &1 &3.38\\
&2 &2.76\\
CFM\cite{yang2024consistency} &2 &5.34&8.70\\
\toprule
\multicolumn{4}{l}{\textbf{Other Generative Models}}\\
PFGM  \cite{pfgm} &110 &2.35 &9.68 \\
PFGM++  \cite{xu2023pfgm++} &110 &2.35 &9.68 \\
\toprule
\multicolumn{4}{l}{\textbf{Integration Flow Models}}\\
\textbf{VE} &1 &2.86 &9.56 \\
&2 &2.62 &9.76 \\


\textbf{1-Rectified Flow} &1 &\textbf{3.36} &9.49\\
&2 &\textbf{2.75} \\

\textbf{PFGM++, D=2048} &1 &\textbf{2.91} & \\

\bottomrule
         
    \end{tabular}
    
    \label{tab:cifar}
\end{table}

\begin{table}[t!]
\setlength\tabcolsep{0.01cm}
    \centering
    \caption{Performance evaluation of class-conditional samples on ImageNet 64x64.}
    \fontsize{8.5pt}{8.5pt}\selectfont
    \begin{tabular}{lcccc}

    \toprule
         Method & NFE($\downarrow$) & FID($\downarrow$) & Prec.($\uparrow$) & Rec.($\uparrow$)\\
         \toprule
         \multicolumn{5}{l}{\textbf{Diffusion Models - Fast Samplers \& Distillation}} \\
         
        DDIM \cite{ddim}&50 &13.7 &0.65 &0.56 \\
        &10 &18.3 &0.60 &0.49 \\
    DPM solver \cite{dpm-solver} &10 &7.93 \\
    &20 &3.42 \\
DEIS \cite{3deis}&10 &6.65\\
&20 &3.10 \\
DFNO (LPIPS) \cite{DFNO} &1 &7.83 &0.61 \\
TRACT \cite{tract}&1 &7.43 \\
&2 &4.97 \\
BOOT \cite{boot}&1 &16.3 &0.68 &0.36 \\
Diff-Instruct \cite{instruct} &1 &5.57\\

PD (LPIPS) \cite{consistency}  &1 &7.88 &0.66 &0.63 \\
&2 &5.74 &0.67 &0.65 \\
         \toprule 
       \multicolumn{5}{l}{\textbf{Diffusion Models - Direct Generation}} \\
RIN \cite{rin} &1000 &1.23 \\
DDPM  \cite{ddpm}&250 &11.0 &0.67 &0.58 \\
iDDPM \cite{iddpm}&250 &2.92 &0.74 &0.62\\
ADM  \cite{ADM}&250 &2.07 &0.74 &0.63 \\
EDM  \cite{edm}&511 &1.36 \\
\toprule
 \multicolumn{4}{l}{\textbf{GAN Models}}\\
BigGAN-deep \cite{bigGAN} &1 &4.06 &0.79 &0.48 \\
\toprule
\multicolumn{4}{l}{\textbf{Consistency Models}}\\
CT (LPIPS) \cite{consistency} &1 &13.0 &0.71 &0.47 \\
&2 &11.1 &0.69 &0.56 \\
iCT \cite{iCT}  &1 &4.02 &0.70 &0.63 \\
&2 &3.20 &0.73 &0.63 \\
iCT-deep \cite{iCT}  &1 &3.25 &0.72 &0.63 \\
&2 &2.77 &0.74 &0.62 \\
CTM + \text {GAN* } \cite{ctm} &1 &1.92& &0.57\\ 
Multistep-CD \cite{heek2024multistep} &1 &4.3 \\
&2 &2.0\\
Multistep-CT\cite{heek2024multistep} &1 &7.2 \\
&2 &2.7\\

\toprule
\multicolumn{4}{l}{\textbf{Rectified Flows}}\\
2-Rectified flow++ \cite{lee2024improving} &1 &4.31\\
&2 &3.64\\
\toprule
\multicolumn{4}{l}{\textbf{Integration Flow Models}}\\
\textbf{VE} &1 &4.09 \\
&2 &3.28\\


\textbf{1-Rectified Flow} &1 &\textbf{4.35} &\\
&2 &\textbf{3.68} &\\

\textbf{PFGM++, D=2048} &1 &\textbf{4.15} & \\
\bottomrule
    \end{tabular}
    
    \label{tab:imagenet}
\end{table}

\section{Discussion}
\label{sec:discuss}

A key achievement of Integration Flow is its ability to solve different ODE-based models using a single framework, addressing a significant challenge that prior models struggled with. For example, two important generative models, the Rectified Flow and diffusion model, are not unified, but Integration Flow can successfully integrate them. This not only simplifies the landscape of ODE-based generative models but also expands their applicability, making them easier to implement across different domains.

Although Integration Flow has achieved the best performance on one-step generation for Rectified Flow and PFGM++, the performance of Integration Flow for VE is still slightly underperformed compared to the current state-of-the-art. We hypothesize that this small performance gap may be attributed to suboptimal hyperparameters in the loss function. Due to limited computation resources, we are not able to search for the best hyperparameter. 

Additionally, we recognize that different noise schedulers can significantly impact the model's performance. The noise scheduling strategy plays a crucial role in the training dynamics and final performance of the model. We plan to investigate more complex schedulers in future work.

Furthermore, same as DDDM, Integration Flow also needs additional memory consumption during training. One solution is to store \(\mathbf{x}_0^{(n)}\) in a buffer or on disk instead of on the GPU. However, this approach will introduce additional overhead during training due to the need to transfer data back to the GPU. We will fix this in our future work.

\section*{Impact Statement}

This paper presents work whose goal is to advance the field
of Machine Learning. There are many potential societal
consequences of our work, none of which we feel must be
specifically highlighted here.

\nocite{langley00}

\bibliography{example_paper}

\begin{thebibliography}{47}
\providecommand{\natexlab}[1]{#1}
\providecommand{\url}[1]{\texttt{#1}}
\expandafter\ifx\csname urlstyle\endcsname\relax
  \providecommand{\doi}[1]{doi: #1}\else
  \providecommand{\doi}{doi: \begingroup \urlstyle{rm}\Url}\fi

\bibitem[Albergo \& Vanden-Eijnden(2022)Albergo and
  Vanden-Eijnden]{albergo2022building}
Albergo, M.~S. and Vanden-Eijnden, E.
\newblock Building normalizing flows with stochastic interpolants.
\newblock \emph{arXiv preprint arXiv:2209.15571}, 2022.

\bibitem[Berthelot et~al.(2023)Berthelot, Autef, Lin, Yap, Zhai, Hu, Zheng,
  Talbot, and Gu]{tract}
Berthelot, D., Autef, A., Lin, J., Yap, D.~A., Zhai, S., Hu, S., Zheng, D.,
  Talbot, W., and Gu, E.
\newblock Tract: Denoising diffusion models with transitive closure
  time-distillation.
\newblock \emph{arXiv preprint arXiv:2303.04248}, 2023.

\bibitem[Bortoli et~al.(2023)Bortoli, Thornton, Heng, and Doucet]{bridge}
Bortoli, V.~D., Thornton, J., Heng, J., and Doucet, A.
\newblock Diffusion schr\"odinger bridge with applications to score-based
  generative modeling, 2023.

\bibitem[Brock et~al.(2018)Brock, Donahue, and Simonyan]{bigGAN}
Brock, A., Donahue, J., and Simonyan, K.
\newblock Large scale gan training for high fidelity natural image synthesis.
\newblock \emph{arXiv preprint arXiv:1809.11096}, 2018.

\bibitem[Charbonnier et~al.(1997)Charbonnier, Blanc-Feraud, Aubert, and
  Barlaud]{PH}
Charbonnier, P., Blanc-Feraud, L., Aubert, G., and Barlaud, M.
\newblock Deterministic edge-preserving regularization in computed imaging.
\newblock \emph{IEEE Transactions on Image Processing}, 6\penalty0
  (2):\penalty0 298--311, 1997.
\newblock \doi{10.1109/83.551699}.

\bibitem[Chen et~al.(2018)Chen, Rubanova, Bettencourt, and
  Duvenaud]{chen2018neural}
Chen, R.~T., Rubanova, Y., Bettencourt, J., and Duvenaud, D.~K.
\newblock Neural ordinary differential equations.
\newblock \emph{Advances in neural information processing systems}, 31, 2018.

\bibitem[Deng et~al.(2009)Deng, Dong, Socher, Li, Li, and Fei-Fei]{imagenet}
Deng, J., Dong, W., Socher, R., Li, L.-J., Li, K., and Fei-Fei, L.
\newblock Imagenet: A large-scale hierarchical image database.
\newblock In \emph{2009 IEEE conference on computer vision and pattern
  recognition}, pp.\  248--255. Ieee, 2009.

\bibitem[Dhariwal \& Nichol(2021)Dhariwal and Nichol]{ADM}
Dhariwal, P. and Nichol, A.
\newblock Diffusion models beat gans on image synthesis.
\newblock \emph{Advances in neural information processing systems},
  34:\penalty0 8780--8794, 2021.

\bibitem[Gholami et~al.(2019)Gholami, Keutzer, and Biros]{gholami2019anode}
Gholami, A., Keutzer, K., and Biros, G.
\newblock Anode: Unconditionally accurate memory-efficient gradients for neural
  odes.
\newblock \emph{arXiv preprint arXiv:1902.10298}, 2019.

\bibitem[Gu et~al.(2023)Gu, Zhai, Zhang, Liu, and Susskind]{boot}
Gu, J., Zhai, S., Zhang, Y., Liu, L., and Susskind, J.
\newblock Boot: Data-free distillation of denoising diffusion models with
  bootstrapping.
\newblock \emph{arXiv preprint arXiv:2306.05544}, 2023.

\bibitem[Heek et~al.(2024)Heek, Hoogeboom, and Salimans]{heek2024multistep}
Heek, J., Hoogeboom, E., and Salimans, T.
\newblock Multistep consistency models.
\newblock \emph{arXiv preprint arXiv:2403.06807}, 2024.

\bibitem[Heusel et~al.(2017)Heusel, Ramsauer, Unterthiner, Nessler, and
  Hochreiter]{FID}
Heusel, M., Ramsauer, H., Unterthiner, T., Nessler, B., and Hochreiter, S.
\newblock Gans trained by a two time-scale update rule converge to a local nash
  equilibrium.
\newblock \emph{Advances in neural information processing systems}, 30, 2017.

\bibitem[Ho \& Salimans(2022)Ho and Salimans]{classifierfree}
Ho, J. and Salimans, T.
\newblock Classifier-free diffusion guidance, 2022.

\bibitem[Ho et~al.(2020)Ho, Jain, and Abbeel]{ddpm}
Ho, J., Jain, A., and Abbeel, P.
\newblock Denoising diffusion probabilistic models.
\newblock \emph{Advances in neural information processing systems},
  33:\penalty0 6840--6851, 2020.

\bibitem[Jabri et~al.(2022)Jabri, Fleet, and Chen]{rin}
Jabri, A., Fleet, D., and Chen, T.
\newblock Scalable adaptive computation for iterative generation.
\newblock \emph{arXiv preprint arXiv:2212.11972}, 2022.

\bibitem[Karras et~al.(2020{\natexlab{a}})Karras, Laine, Aittala, Hellsten,
  Lehtinen, and Aila]{style2}
Karras, T., Laine, S., Aittala, M., Hellsten, J., Lehtinen, J., and Aila, T.
\newblock Analyzing and improving the image quality of stylegan.
\newblock In \emph{Proceedings of the IEEE/CVF conference on computer vision
  and pattern recognition}, pp.\  8110--8119, 2020{\natexlab{a}}.

\bibitem[Karras et~al.(2020{\natexlab{b}})Karras, Laine, Aittala, Hellsten,
  Lehtinen, and Aila]{stylegan2-ada}
Karras, T., Laine, S., Aittala, M., Hellsten, J., Lehtinen, J., and Aila, T.
\newblock Analyzing and improving the image quality of stylegan.
\newblock In \emph{Proceedings of the IEEE/CVF conference on computer vision
  and pattern recognition}, pp.\  8110--8119, 2020{\natexlab{b}}.

\bibitem[Karras et~al.(2022)Karras, Aittala, Aila, and Laine]{edm}
Karras, T., Aittala, M., Aila, T., and Laine, S.
\newblock Elucidating the design space of diffusion-based generative models.
\newblock \emph{Advances in Neural Information Processing Systems},
  35:\penalty0 26565--26577, 2022.

\bibitem[Kim et~al.(2022)Kim, Kim, Kwon, Kang, and Moon]{edmG}
Kim, D., Kim, Y., Kwon, S.~J., Kang, W., and Moon, I.-C.
\newblock Refining generative process with discriminator guidance in
  score-based diffusion models.
\newblock \emph{arXiv preprint arXiv:2211.17091}, 2022.

\bibitem[Kim et~al.(2024)Kim, Lai, Liao, Murata, Takida, Uesaka, He, Mitsufuji,
  and Ermon]{ctm}
Kim, D., Lai, C.-H., Liao, W.-H., Murata, N., Takida, Y., Uesaka, T., He, Y.,
  Mitsufuji, Y., and Ermon, S.
\newblock Consistency trajectory models: Learning probability flow ode
  trajectory of diffusion, 2024.

\bibitem[Kong et~al.(2021)Kong, Ping, Huang, Zhao, and
  Catanzaro]{kong2021diffwave}
Kong, Z., Ping, W., Huang, J., Zhao, K., and Catanzaro, B.
\newblock Diffwave: A versatile diffusion model for audio synthesis.
\newblock In \emph{International Conference on Learning Representations}, 2021.
\newblock URL \url{https://openreview.net/forum?id=a-xFK8Ymz5J}.

\bibitem[Krizhevsky et~al.(2009)Krizhevsky, Hinton, et~al.]{CIFAR}
Krizhevsky, A., Hinton, G., et~al.
\newblock Learning multiple layers of features from tiny images.
\newblock 2009.

\bibitem[Lee et~al.(2024)Lee, Lin, and Fanti]{lee2024improving}
Lee, S., Lin, Z., and Fanti, G.
\newblock Improving the training of rectified flows.
\newblock \emph{arXiv preprint arXiv:2405.20320}, 2024.

\bibitem[Lipman et~al.(2022)Lipman, Chen, Ben-Hamu, Nickel, and
  Le]{lipman2022flow}
Lipman, Y., Chen, R.~T., Ben-Hamu, H., Nickel, M., and Le, M.
\newblock Flow matching for generative modeling.
\newblock \emph{arXiv preprint arXiv:2210.02747}, 2022.

\bibitem[Liu et~al.(2022)Liu, Gong, and Liu]{rectified}
Liu, X., Gong, C., and Liu, Q.
\newblock Flow straight and fast: Learning to generate and transfer data with
  rectified flow.
\newblock \emph{arXiv preprint arXiv:2209.03003}, 2022.

\bibitem[Liu et~al.(2023)Liu, Zhang, Ma, Peng, et~al.]{liu2023instaflow}
Liu, X., Zhang, X., Ma, J., Peng, J., et~al.
\newblock Instaflow: One step is enough for high-quality diffusion-based
  text-to-image generation.
\newblock In \emph{The Twelfth International Conference on Learning
  Representations}, 2023.

\bibitem[Lu et~al.(2022)Lu, Zhou, Bao, Chen, Li, and Zhu]{dpm-solver}
Lu, C., Zhou, Y., Bao, F., Chen, J., Li, C., and Zhu, J.
\newblock Dpm-solver: A fast ode solver for diffusion probabilistic model
  sampling in around 10 steps.
\newblock \emph{Advances in Neural Information Processing Systems},
  35:\penalty0 5775--5787, 2022.

\bibitem[Luhman \& Luhman(2021)Luhman and Luhman]{knowledge}
Luhman, E. and Luhman, T.
\newblock Knowledge distillation in iterative generative models for improved
  sampling speed.
\newblock \emph{arXiv preprint arXiv:2101.02388}, 2021.

\bibitem[Luo et~al.(2023)Luo, Hu, Zhang, Sun, Li, and Zhang]{instruct}
Luo, W., Hu, T., Zhang, S., Sun, J., Li, Z., and Zhang, Z.
\newblock Diff-instruct: A universal approach for transferring knowledge from
  pre-trained diffusion models.
\newblock \emph{arXiv preprint arXiv:2305.18455}, 2023.

\bibitem[Nichol \& Dhariwal(2021)Nichol and Dhariwal]{iddpm}
Nichol, A.~Q. and Dhariwal, P.
\newblock Improved denoising diffusion probabilistic models.
\newblock In \emph{International Conference on Machine Learning}, pp.\
  8162--8171. PMLR, 2021.

\bibitem[Popov et~al.(2022)Popov, Vovk, Gogoryan, Sadekova, Kudinov, and
  Wei]{popov2022diffusionbased}
Popov, V., Vovk, I., Gogoryan, V., Sadekova, T., Kudinov, M., and Wei, J.
\newblock Diffusion-based voice conversion with fast maximum likelihood
  sampling scheme, 2022.

\bibitem[Rombach et~al.(2022)Rombach, Blattmann, Lorenz, Esser, and
  Ommer]{latentDiff}
Rombach, R., Blattmann, A., Lorenz, D., Esser, P., and Ommer, B.
\newblock High-resolution image synthesis with latent diffusion models.
\newblock In \emph{Proceedings of the IEEE/CVF conference on computer vision
  and pattern recognition}, pp.\  10684--10695, 2022.

\bibitem[Saharia et~al.(2022)Saharia, Chan, Saxena, Li, Whang, Denton,
  Ghasemipour, Gontijo~Lopes, Karagol~Ayan, Salimans, et~al.]{Imagen}
Saharia, C., Chan, W., Saxena, S., Li, L., Whang, J., Denton, E.~L.,
  Ghasemipour, K., Gontijo~Lopes, R., Karagol~Ayan, B., Salimans, T., et~al.
\newblock Photorealistic text-to-image diffusion models with deep language
  understanding.
\newblock \emph{Advances in Neural Information Processing Systems},
  35:\penalty0 36479--36494, 2022.

\bibitem[Sauer et~al.(2023)Sauer, Karras, Laine, Geiger, and
  Aila]{sauer2023stylegantunlockingpowergans}
Sauer, A., Karras, T., Laine, S., Geiger, A., and Aila, T.
\newblock Stylegan-t: Unlocking the power of gans for fast large-scale
  text-to-image synthesis, 2023.
\newblock URL \url{https://arxiv.org/abs/2301.09515}.

\bibitem[Song et~al.(2020{\natexlab{a}})Song, Meng, and Ermon]{ddim}
Song, J., Meng, C., and Ermon, S.
\newblock Denoising diffusion implicit models.
\newblock \emph{arXiv preprint arXiv:2010.02502}, 2020{\natexlab{a}}.

\bibitem[Song \& Dhariwal(2023)Song and Dhariwal]{iCT}
Song, Y. and Dhariwal, P.
\newblock Improved techniques for training consistency models.
\newblock \emph{arXiv preprint arXiv:2310.14189}, 2023.

\bibitem[Song et~al.(2020{\natexlab{b}})Song, Sohl-Dickstein, Kingma, Kumar,
  Ermon, and Poole]{score}
Song, Y., Sohl-Dickstein, J., Kingma, D.~P., Kumar, A., Ermon, S., and Poole,
  B.
\newblock Score-based generative modeling through stochastic differential
  equations.
\newblock \emph{arXiv preprint arXiv:2011.13456}, 2020{\natexlab{b}}.

\bibitem[Song et~al.(2023)Song, Dhariwal, Chen, and Sutskever]{consistency}
Song, Y., Dhariwal, P., Chen, M., and Sutskever, I.
\newblock Consistency models.
\newblock \emph{arXiv preprint arXiv:2303.01469}, 2023.

\bibitem[Srivastava et~al.(2014)Srivastava, Hinton, Krizhevsky, Sutskever, and
  Salakhutdinov]{dropout}
Srivastava, N., Hinton, G., Krizhevsky, A., Sutskever, I., and Salakhutdinov,
  R.
\newblock Dropout: a simple way to prevent neural networks from overfitting.
\newblock \emph{The journal of machine learning research}, 15\penalty0
  (1):\penalty0 1929--1958, 2014.

\bibitem[Vahdat et~al.(2021)Vahdat, Kreis, and Kautz]{LSGM}
Vahdat, A., Kreis, K., and Kautz, J.
\newblock Score-based generative modeling in latent space.
\newblock \emph{Advances in Neural Information Processing Systems},
  34:\penalty0 11287--11302, 2021.

\bibitem[Xu et~al.(2022)Xu, Liu, Tegmark, and Jaakkola]{pfgm}
Xu, Y., Liu, Z., Tegmark, M., and Jaakkola, T.
\newblock Poisson flow generative models.
\newblock \emph{Advances in Neural Information Processing Systems},
  35:\penalty0 16782--16795, 2022.

\bibitem[Xu et~al.(2023)Xu, Liu, Tian, Tong, Tegmark, and
  Jaakkola]{xu2023pfgm++}
Xu, Y., Liu, Z., Tian, Y., Tong, S., Tegmark, M., and Jaakkola, T.
\newblock Pfgm++: Unlocking the potential of physics-inspired generative
  models.
\newblock In \emph{International Conference on Machine Learning}, pp.\
  38566--38591. PMLR, 2023.

\bibitem[Yang et~al.(2024)Yang, Zhang, Zhang, Liu, Xu, Zhang, Meng, Ermon, and
  Cui]{yang2024consistency}
Yang, L., Zhang, Z., Zhang, Z., Liu, X., Xu, M., Zhang, W., Meng, C., Ermon,
  S., and Cui, B.
\newblock Consistency flow matching: Defining straight flows with velocity
  consistency.
\newblock \emph{arXiv preprint arXiv:2407.02398}, 2024.

\bibitem[Zhang et~al.(2024)Zhang, Wang, and Luo]{zhang2024directly}
Zhang, D., Wang, J., and Luo, F.
\newblock Directly denoising diffusion model.
\newblock \emph{arXiv preprint arXiv:2405.13540}, 2024.

\bibitem[Zhang \& Chen(2022)Zhang and Chen]{3deis}
Zhang, Q. and Chen, Y.
\newblock Fast sampling of diffusion models with exponential integrator.
\newblock \emph{arXiv preprint arXiv:2204.13902}, 2022.

\bibitem[Zhao et~al.(2023)Zhao, Bai, Rao, Zhou, and Lu]{unipc}
Zhao, W., Bai, L., Rao, Y., Zhou, J., and Lu, J.
\newblock Unipc: A unified predictor-corrector framework for fast sampling of
  diffusion models.
\newblock \emph{arXiv preprint arXiv:2302.04867}, 2023.

\bibitem[Zheng et~al.(2023)Zheng, Nie, Vahdat, Azizzadenesheli, and
  Anandkumar]{DFNO}
Zheng, H., Nie, W., Vahdat, A., Azizzadenesheli, K., and Anandkumar, A.
\newblock Fast sampling of diffusion models via operator learning.
\newblock In \emph{International Conference on Machine Learning}, pp.\
  42390--42402. PMLR, 2023.

\end{thebibliography}
\bibliographystyle{icml2025}

\newpage
\appendix
\onecolumn

\section{Derivation of Integration Flow Algorithms}\label{App}

\subsection{Integration Flow for VE diffusion models}\label{App1}

The PF-ODE of VE diffusion models is formulated as: 
\begin{equation}
\frac{\mathrm{d} \mathbf{x}_t}{\mathrm{~d} t}=-\frac{1}{2} \frac{\mathrm{~d} \sigma_t^2}{\mathrm{~d} t} \nabla_{\mathbf{x}_t} \log p_t\left(\mathbf{x}_t\right)
\end{equation}
We do the reversed time integration on both sides over the interval $[0, t]$:

\begin{equation}
\int_t^0 \frac{\mathrm{d} \mathbf{x}_s}{\mathrm{~d} s}=\int_t^0-\frac{1}{2} \frac{\mathrm{~d} \sigma_t^2}{\mathrm{~d} t} \nabla_{\mathbf{x}_t} \log p_t\left(\mathbf{x}_t\right) \end{equation}

and obtain:

\begin{equation}
    \mathbf{x}_0-\mathbf{x}_t=V\left(\mathbf{x}_0, 0\right)-V\left(\mathbf{x}_t, t\right)=-G\left(\mathbf{x}_0, \mathbf{x}_t, t\right)
    \end{equation}

Thus:
\begin{equation}
\mathbf{x}_0 =\mathbf{x}_t-G\left(\mathbf{x}_0, \mathbf{x}_t, t\right) = g\left(\mathbf{x}_0, \mathbf{x}_t, t\right) 
\end{equation}

For stable training purpose, we rewrite $g\left(\mathbf{x}_0, \mathbf{x}_t, t\right) $ as following: 

\begin{equation}
g\left(\mathbf{x}_0, \mathbf{x}_t, t\right) = \mathbf{x}_t - G\left(\mathbf{x}_0, \mathbf{x}_t, t\right) = \kappa\left(\sigma_t\right) \mathbf{x}_t + \left(1 - \kappa\left(\sigma_t\right)\right) *\left[\mathbf{x}_t - \frac{1}{1 - \kappa\left(\sigma_t\right)} G\left(\mathbf{x}_0, \mathbf{x}_t, t\right)\right]
\end{equation}
We define the neural network as: 
\begin{equation}
g_{\boldsymbol{\theta}}\left(\mathbf{x}_0, \mathbf{x}_t, t\right) = \kappa\left(\sigma_t\right) \mathbf{x}_t+
\frac{1}{1-\kappa\left(\sigma_t\right)} G_{\boldsymbol{\theta}}\left(\mathbf{x}_0, \mathbf{x}_t, t\right),
\end{equation}
where we use neural network to estimate the value of $\mathbf{x}_t-\left(1-\kappa\left(\sigma_t\right)\right)G\left(\mathbf{x}_0, \mathbf{x}_t, t\right)$.

For VE Diffusion model, we have $a(\sigma_t)=\kappa(\sigma_t)$ and  $b(\sigma_t)=1-\kappa(\sigma_t)$.
There are a few choices for the design of \(\kappa(\sigma_t)\), such as \(\kappa(\sigma_t) = \frac{\sigma_{\text{data}}}{\sigma_t+\sigma_{\text{data}}}\), \(\kappa(\sigma_t) = \frac{\sigma_{\text{data}}^2}{\sigma_t^2+\sigma_{\text{data}}^2}\), which are in a manner of \citep{edm}. We set \(\kappa(\sigma_t) = \frac{\sigma_{\text{min}}}{\sigma_t}\) in this work.

\begin{algorithm}[ht]
   \caption{Integration Flow Training Algorithm for VE Diffusion Model}
\begin{algorithmic}
   \STATE {\bfseries Input:} $p_{\text{data}}, T$, model parameter $\boldsymbol\theta$, initialize $\mathbf{x}_0^{(0)} \sim \mathcal{N}(\mathbf{0}, \mathbf{I})$,
    epoch $n \leftarrow 0$
   \REPEAT
   \STATE Sample $\mathbf{x}_0 \sim p_{\text{data}}$, $t \sim \mathcal{U}\left[1, T\right]$ and $\boldsymbol{\epsilon} \sim \mathcal{N}(\mathbf{0}, \mathbf{I})$
   
   \STATE $\mathbf{x}_t = \mathbf{x}_0+\sigma_t\boldsymbol{\epsilon}$
  
   \STATE$\mathbf{x}_0^{(n+1)} \leftarrow g_{\boldsymbol\theta}\left(\mathbf{x}_{0}^{(n)}, \mathbf{x}_t, t\right)$ 
    \STATE $\mathcal{L}_{\text{IFM}}^{(n+1)}(\boldsymbol{\theta})\leftarrow d\left(\mathbf{x}_0^{(n+1)}, \mathbf{x}_0\right)$
   
   \STATE $\boldsymbol{\theta} \leftarrow \boldsymbol{\theta}-\eta \nabla_{\boldsymbol{\theta}} \mathcal{L}\left(\boldsymbol{\theta}\right)$
   \STATE $n \leftarrow n+1$
   \UNTIL{convergence}
\end{algorithmic}
\label{algo:1}
\end{algorithm}

\begin{algorithm}[ht]
   \caption{ Integration Flow Sampling Algorithm for VE Diffusion Model}
\begin{algorithmic}
   \STATE {\bfseries Input:} $T$, trained model parameter $\boldsymbol\theta$, sampling step $k$, initialize
   $\mathbf{x}_0^{(0)} \sim \mathcal{N}(\mathbf{0}, \mathbf{I})$, $\mathbf{x}_T \sim \mathcal{N}(\mathbf{0}, \mathbf{I})$
   
   \STATE $\mathbf{x}_T = \sigma_\text{max}\mathbf{x}_T$
   
   \FOR{$k=0$ {\bfseries to} $k-1$}
   \STATE $\mathbf{x}_0^{(k+1)} \leftarrow g_{\boldsymbol\theta}\left(\mathbf{x}_{0}^{(k)}, \mathbf{x}_T, T\right)$ 
   \ENDFOR
   \STATE \textbf{Output:} $\mathbf{x}_0^{(k+1)}$
   
\end{algorithmic}
\label{algo:2}
\end{algorithm}

\subsection{Integration Flow for Rectified Flow}\label{App2}




To analyze the integration flow associated with this process, we consider the derivative of $\mathbf{z}_t$ with respect to time $t \in [0,1]$:

$$\frac{d \mathbf{z}_t}{d t}=v\left(\mathbf{z}_t, t\right)$$


Integrating the both side, we obtain:

$$ \int_0^1 \frac{d \mathbf{z}_s}{d s} ds = \mathbf{z} - \mathbf{x}_0. $$

This confirms that the total change over the entire path from $t = 0$ to $t = 1$ is simply the difference between the endpoints $\mathbf{z}$ and $\mathbf{x}_0$.

For any intermediate time $t \in [0, 1]$, reserve-time integration over $[0, t]$ yields:

\begin{equation}
    \int_t^0 \frac{d \mathbf{z}_s}{d s} ds = \mathbf{x}_0 - \mathbf{z}_t = V\left(\mathbf{x}_0, 0\right)-V\left(\mathbf{x}_t, t\right)= - G(\mathbf{x}_0, \mathbf{z}_t, t),  
    \end{equation} 

where we define the accumulated change $G(\mathbf{x}_0, \mathbf{z}_t, t)$ as:

\begin{equation}
 G(\mathbf{x}_0, \mathbf{z}_t, t) = \mathbf{z}_t - \mathbf{x}_0. \end{equation}

Substituting the expression for $\mathbf{z}_t$, we have:

\begin{equation}
\mathbf{z}_t - \mathbf{x}_0 = [(1 - t) \mathbf{x}_0 + t \mathbf{z}] - \mathbf{x}_0 = t (\mathbf{z} - \mathbf{x}_0). \end{equation}

Thus, the accumulated change is proportional to the time parameter $t$ and the difference $\mathbf{z} - \mathbf{x}_0$:

\begin{equation}
G(\mathbf{x}_0, \mathbf{z}_t, t) = t (\mathbf{z} - \mathbf{x}_0). 
\end{equation}

Rearranging this expression allows us to solve for $\mathbf{z} - \mathbf{x}_0$:

\begin{equation} \mathbf{z}- \mathbf{x}_0 = \frac{G(\mathbf{x}_0, \mathbf{z}_t, t)}{t}. \end{equation}

This relationship indicates that the total accumulated change $\mathbf{z} - \mathbf{x}_0$ can be expressed in terms of the accumulated change $G(\mathbf{x}_0, \mathbf{z}_t, t)$ scaled by $1/t$.

We can now define the Integration Flow of the Rectified Flow process by expressing $\mathbf{x}_0$ in terms of $G(\mathbf{x}_0, \mathbf{z}_t, t)$ and the endpoint $\mathbf{z}$:

\begin{equation}
    \mathbf{x}_0 = \mathbf{z} - \frac{G(\mathbf{x}_0, \mathbf{z}_t, t)}{t} = g\left(\mathbf{x}_0, \mathbf{x}_t, t\right).
\end{equation}

In practice, since $\mathbf{z}$ is deterministic, it can be absorbed into the neural network; for stable training, we take $G\left(\mathbf{x}_0, \mathbf{z}_t, t\right)/t$ as a whole. 

Thus ,we have:

$$
g_{\boldsymbol{\theta}}\left(\mathbf{x}_0, \mathbf{x}_t, t\right)=G_{\boldsymbol{\theta}}\left(\mathbf{x}_0, \mathbf{x}_t, t\right),
$$

which indicates $a_t = 0$ and $b_t = 1$.

By employing this enhanced dynamic model within the Rectified Flow framework, we can achieve a more accurate and stable reconstruction of the initial data point $\mathbf{x}_0$, facilitating effective generative modeling and data synthesis.

\begin{algorithm}[ht]

   \caption{Integration Flow Training Algorithm for Rectified Flows}
\begin{algorithmic}
   \STATE {\bfseries Input:} couple $\left(\mathbf{x}_0, \mathbf{z}\right)$ from $p_{\text{data}}$ and $p_{\mathbf{z}},$ respectively; model parameter $\boldsymbol\theta$, initialize $\mathbf{x}_0^{(0)} \sim \mathcal{N}(\mathbf{0}, \mathbf{I})$, epoch $n \leftarrow 0$ 
   

     \REPEAT
   \STATE Sample $\mathbf{x}_0 \sim p_{\text{data}}, t\sim \text{Uniform}[0,1]$ 
   
   \STATE $\mathbf{z}_t=(1-t) \mathbf{x}_0+t \mathbf{z}$
  
   \STATE$\mathbf{x}_0^{(n+1)} \leftarrow g_{\boldsymbol\theta}\left(\mathbf{x}_{0}^{(n)}, \mathbf{z}_t, t\right)$ 
    \STATE $\mathcal{L}_{\text{IFM}}^{(n+1)}(\boldsymbol{\theta})\leftarrow d\left(\mathbf{x}_0^{(n+1)}, \mathbf{x}_0\right)$

   \STATE $\boldsymbol{\theta} \leftarrow \boldsymbol{\theta}-\eta \nabla_{\boldsymbol{\theta}} \mathcal{L}\left(\boldsymbol{\theta}\right)$
   \STATE $n \leftarrow n+1$
   \UNTIL{convergence}

\end{algorithmic}
\label{algo:3}
\end{algorithm}

\begin{algorithm}[ht]
   \caption{Integration Flow Sampling Algorithm for Rectified Flows}
\begin{algorithmic}
   \STATE {\bfseries Input:} $t = 1,$ trained model parameter $\boldsymbol\theta$, draw $\mathbf{z} \sim p_{\mathbf{z}}$, initialize $\mathbf{x}_0^{(0)} \sim \mathcal{N}(\mathbf{0}, \mathbf{I}),$ sampling step $k,$ initialize
   $\mathbf{x}_0^{(0)} \sim \mathcal{N}(\mathbf{0}, \mathbf{I})$
   \FOR{$k=0$ {\bfseries to} $k-1$}
   \STATE $\mathbf{x}_0^{(k+1)} \leftarrow g_{\boldsymbol\theta}\left(\mathbf{x}_{0}^{(k)}, \mathbf{z}, t\right)$ 
   \ENDFOR
   \STATE \textbf{Output:} $\mathbf{x}_0^{(k+1)}$
   
   
\end{algorithmic}
\label{algo:4}
\end{algorithm}

\subsection{Integration Flow for PFGM++}\label{App3}

The backward ODE of PFGM++ is characterized as:

\begin{equation}
\frac{d \mathbf{x}}{d r}=\frac{\mathbf{E}(\tilde{\mathbf{x}})_{\mathbf{x}}}{E(\tilde{\mathbf{x}})_r}\label{eq 8}
\end{equation}
Since \citep{xu2023pfgm++} showed that $d \mathbf{x}/d r =d \mathbf{x}/d \sigma$,where $\sigma$ changes with time.

We modify the \eqref{eq 8} as:
\begin{equation}
\frac{d \mathbf{x}}{d t}=\frac{d \mathbf{x}}{d r}\frac{d \sigma_t}{d t} =\frac{\mathbf{E}(\tilde{\mathbf{x}})_{\mathbf{x}}}{E(\tilde{\mathbf{x}})_r}\frac{d \sigma_t}{d t}
\end{equation}
We do the reversed time integration on both sides with respect to $t$ over the interval $[0, t]$, leading to:
\begin{equation}
 \int_t^0 \frac{d \mathbf{x}}{d t}  dt = \int_t^0 \frac{\mathbf{E}(\tilde{\mathbf{x}})_{\mathbf{x}}}{E(\tilde{\mathbf{x}})_r} \frac{d \sigma_t}{d t} d t, \end{equation}
which is equivalent to:
\begin{equation}
    \mathbf{x}_0 - \mathbf{x}_t = V\left(\mathbf{x}_0, 0\right)-V\left(\mathbf{x}_t, t\right) = - G(\mathbf{x}_0, \mathbf{x}_t, t)  
\end{equation}

Rearranging the equation, we express the initial data point in terms of the integration flow:

$$\mathbf{x}_0 = \mathbf{x}_t  - G(\mathbf{x}_0, \mathbf{x}_t, t) =g\left(\mathbf{x}_0, \mathbf{x}_t, t\right). $$

For stable training purpose, we rewrite $g\left(\mathbf{x}_0, \mathbf{x}_t, t\right) $ as following: 
\begin{equation}
g\left(\mathbf{x}_0, \mathbf{x}_t, t\right) =\mathbf{x}_t- G\left(\mathbf{x}_0, \mathbf{x}_t, t\right) = a_t\mathbf{x}_t+(1-a_t)\left[\mathbf{x}_t-\frac{1}{1-a_t} G\left(\mathbf{x}_0, \mathbf{x}_t, t\right)\right]
\end{equation}

Since $\mathbf{x}_t=\mathbf{x}_0+R_t \mathbf{v}_t$, inspired by the settings of $a_t, b_t$ in VE case of diffusion model, we set $a_t = \sigma_{\min }/R_t,  b_t = 1-a_t = 1-\sigma_{\min }/R_t.$


\begin{algorithm}[H]
   \caption{Integration Flow Training Algorithm for PFGM++}
\begin{algorithmic}
   \STATE {\bfseries Input:} $p_{\text{data}}, T$, model parameter $\boldsymbol\theta$, initialize $\mathbf{x}_0^{(0)} \sim \mathcal{N}(\mathbf{0}, \mathbf{I})$,
    epoch $n \leftarrow 0$
   \REPEAT 
   \STATE Sample $\mathbf{x}_0 \sim p_{\text{data}}, t \sim \mathcal{U}\left[1, T\right]$
   \STATE Sample $R_t, \mathbf{v}_t$ where:
          \STATE \quad $R_t = \sigma_t \sqrt{D}, \quad R_t \sim p_{r_t}(R)$
          \STATE \quad $\mathbf{v}_t = \frac{\mathbf{u}_t}{\left\|\mathbf{u}_t\right\|_2}, \quad \mathbf{u}_t \sim \mathcal{N}(\mathbf{0}, \mathbf{I})$
           \STATE Compute $\mathbf{x}_t = \mathbf{x}_0 + R_t \mathbf{v}_t$

   \STATE$\mathbf{x}_0^{(n+1)} \leftarrow g_{\boldsymbol\theta}\left(\mathbf{x}_{0}^{(n)}, \mathbf{x}_t, t\right)$ 
    \STATE $\mathcal{L}_{\text{IFM}}^{(n+1)}(\boldsymbol{\theta})\leftarrow d\left(\mathbf{x}_0^{(n+1)}, \mathbf{x}_0\right)$
   
   \STATE $\boldsymbol{\theta} \leftarrow \boldsymbol{\theta}-\eta \nabla_{\boldsymbol{\theta}} \mathcal{L}\left(\boldsymbol{\theta}\right)$
   \STATE $n \leftarrow n+1$
   \UNTIL{convergence}
\end{algorithmic}
\label{algo:5}
\end{algorithm}

\begin{algorithm}[h]
   \caption{Integration Flow Sampling Algorithm for PFGM++}
\begin{algorithmic}
   \STATE {\bfseries Input:} $T$, trained model parameter $\boldsymbol\theta$, sampling step $k$, $r_{T }=\sigma_{T} \sqrt{D}, R_T \sim p_{r_{T}}(R), \mathbf{v}=\frac{\mathbf{u}}{\|\mathbf{u}\|_2},$ with $\mathbf{u} \sim \mathcal{N}(\mathbf{0}, \mathbf{I})$, initialize
   $\mathbf{x}_0^{(0)} \sim \mathcal{N}(\mathbf{0}, \mathbf{I})$
   
   \STATE Compute $\mathbf{x}_T = R_T \mathbf{v}$
   
   \FOR{$k=0$ {\bfseries to} $k-1$}
   \STATE $\mathbf{x}_0^{(k+1)} \leftarrow g_{\boldsymbol\theta}\left(\mathbf{x}_{0}^{(k)}, \mathbf{x}_T, T\right)$ 
   \ENDFOR
   \STATE \textbf{Output:} $\mathbf{x}_0^{(n+1)}$
\end{algorithmic}
\label{algo:6}
\end{algorithm}

\section{Proof of Theorems}\label{AppB}
\subsection {Proof of \textbf{Theorem 1}:}
\begin{proof}
Before proving the theorem, we prove the corollary first:

\textbf{Corollary}: 
\begin{equation}
\mathbb{E}\left[\left(A-\mathbb{E}[A \mid B, C]\right)^2\right] \leq \mathbb{E}\left[\left(A-\mathbb{E}[A \mid B]\right)^2\right],
\end{equation}

The variance of a random variable $A$ can be decomposed as follows:
\begin{equation}
\operatorname{Var}(A)=\mathbb{E}[\operatorname{Var}(A \mid B)]+\operatorname{Var}(\mathbb{E}[A \mid B]).
\end{equation}
Now, introduce the extra information $C$. The variance of $A$, conditioned on both $B$ and $C$, is:

\begin{equation}
\operatorname{Var}(A|B, C)=\mathbb{E}[\operatorname{Var}(A|B, C) | B]+\operatorname{Var}(\mathbb{E}[A |B, C]| B)
\end{equation}

Since conditioning on more information reduces uncertainty, we know that:

\begin{equation}
\operatorname{Var}(A|B, C) \leq \operatorname{Var}(A|B).
\end{equation}

We also have:
\begin{equation}
\mathbb{E}\left[\left(A-\mathbb{E}[A|B, C]\right)^2\right]=\mathbb{E}[\operatorname{Var}(A|B, C)]
\end{equation}
and
\begin{equation}
\mathbb{E}\left[\left(A-\mathbb{E}[A|B]\right)^2\right]=\mathbb{E}[\operatorname{Var}(A|B)]
\end{equation}

 Since $\operatorname{Var}(A|B, C) \leq \operatorname{Var}(A| B)$, it follows that:

\begin{equation}
\mathbb{E}\left[\left(A-\mathbb{E}[A|B, C]\right)^2\right] \leq \mathbb{E}\left[\left(A-\mathbb{E}[A|B]\right)^2\right]
\end{equation}

let $\mathbf{x}_0 = A, \mathbf{x}_t = B, \mathbf{x}_0^{(n)}= C$, plug into the corollary, we complete the proof
\end{proof}

\subsection {Proof of \textbf{Theorem 2}:}

\begin{proof}

The initial value problem (IVP) of the reversed time ODE can be expressed as:

\begin{equation}
\label{ivp:1}
\left\{
\begin{aligned}
    \frac{\mathrm{d} \mathbf{x}_s}{\mathrm{~d} s}&=v\left( \mathbf{x}_s, s\right) && \quad s \in [0, t] \\
    \mathbf{x}_t &= \hat{\mathbf{x}}_t
\end{aligned}
\right.
\end{equation}

if putting \(\Tilde{\mathbf{x}}_s:=\mathbf{x}_{t-s}\), we get 
\begin{equation}
\label{ivp:2}
\left\{
\begin{aligned}
    \frac{\mathrm{~d} \Tilde{\mathbf{x}}_s}{\mathrm{~d} s}&=-v\left( \Tilde{\mathbf{x}}_s, s\right) && \quad s \in [0, t] \\
    \Tilde{\mathbf{x}}_0 &= \hat{\mathbf{x}}_t
\end{aligned}
\right.
\end{equation}

The IVP \ref{ivp:1} and \ref{ivp:2} are equivalent and can be used interchangeably.

From the Lipschitz condition on $v$, we have:
\begin{equation}
\left\|v( \tilde{\mathbf{x}}_s, s)-v(\tilde{\mathbf{y}}_s, s)\right\|_2 \leq L\left\|\tilde{\mathbf{x}}_s-\tilde{\mathbf{y}}_s\right\|_2 .
\end{equation}

Use the integral form:
\begin{equation}
\left\|g\left(\mathbf{x}_0, \mathbf{x}_t, t\right)-g\left(\mathbf{y}_0, \mathbf{y}_t, t\right)\right\|_2 \leq \left\|\tilde{\mathbf{x}}_0-\tilde{\mathbf{y}}_0\right\|_2 + \int_0^t L\left\|\tilde{\mathbf{x}}_s-\tilde{\mathbf{y}}_s\right\|_2  d s
\end{equation}

By using Gröwnwall inequality, we have:
\begin{equation}
\left\|g\left(\mathbf{x}_0, \mathbf{x}_t, t\right)-g\left(\mathbf{y}_0, \mathbf{y}_t, t\right)\right\|_2   \leq e^{Lt}\left\|\tilde{\mathbf{x}}_0-\tilde{\mathbf{y}}_0\right\|_2 = e^{Lt}\left\|\hat{\mathbf{x}}_t-\hat{\mathbf{y}}_t\right\|_2
\end{equation}
Next, consider the inverse time ODE, we have: 
\begin{equation}
\left\|\mathbf{x}_t-\mathbf{y}_t\right\|_2 \leq \left\|g(\mathbf{x}_0, \mathbf{x}_t, t)-g(\mathbf{y}_0, \mathbf{y}_t, t)\right\|_2 +\int_0^t L\left\|\mathbf{x}_s-\mathbf{y}_s\right\|_2 d s
\end{equation}
Again, by using Gröwnwall inequality,
\begin{equation}
\left\|\mathbf{x}_t-\mathbf{y}_t\right\|_2 \leq e^{Lt}\left\|g(\mathbf{x}_0, \mathbf{x}_t, t)-g(\mathbf{y}_0, \mathbf{y}_t, t)\right\|_2 
\end{equation}
Therefore,
\begin{equation}
\left\|g\left(\mathbf{x}_0, \mathbf{x}_t, t\right)-g\left(\mathbf{y}_0, \mathbf{y}_t, t\right)\right\|_2 \geq e^{-L t}\left\|\mathbf{x}_t-\mathbf{y}_t\right\|_2 
\end{equation}
and we complete the proof of:
\begin{equation}
\begin{split}
e^{-Lt}\left\|\mathbf{x}_t-\mathbf{y}_t\right\|_2 \leq \left\|g(\mathbf{x}_0,\mathbf{x}_t, t)-g(\mathbf{y}_0,\mathbf{y}_t, t)\right\|_2 \leq e^{Lt}\left\|\mathbf{x}_t-\mathbf{y}_t\right\|_2 . \label{eq 11}
\end{split}
\end{equation}

Since the neural network is sufficiently trained, we have:

$
g_{\boldsymbol{\theta}^*}(\mathbf{x}_0^{(n)}, \mathbf{x}_t, t) \equiv g(\mathbf{x}_0, \mathbf{x}_t, t)$, replace $g(\mathbf{x}_0, \mathbf{x}_t, t)$, $g(\mathbf{y}_0, \mathbf{y}_t, t)$ with $g_{\boldsymbol{\theta}^*}(\mathbf{x}_0^{(n)}, \mathbf{x}_t, t)$, $g_{\boldsymbol{\theta}^*}(\mathbf{y}_0^{(n)}, \mathbf{y}_t, t)$ respectively in \eqref{eq 11}, we obtain:
\begin{equation}
\begin{split}
e^{-L t}\left\|\mathbf{x}_t-\mathbf{y}_t\right\|_2 \leq\left\|g_{\boldsymbol{\theta}^*}(\mathbf{x}_0^{(n)}, \mathbf{x}_t, t)-g_{\boldsymbol{\theta}^*}(\mathbf{y}_0^{(n)}, \mathbf{y}_t, t)\right\|_2 \leq e^{L t}\left\|\mathbf{x}_t-\mathbf{y}_t\right\|_2
\end{split}
\end{equation}

\end{proof}

\subsection {Proof of \textbf{Theorem 3}:}

\begin{proof} For clarity within this particular theorem, we adopt the same notational convention as rectified flow: we label the initial (starting) points by $\mathbf{x}_0$ and the final (target) points by $\mathbf{x}_1$. Everywhere else in the paper (outside this theorem), we follow the convention of diffusion model, where $\mathbf{x}_0$ denotes the sample to be recovered (the "target" in the diffusion context), and $\mathbf{x}_1$ or $\mathbf{x}_T$ denotes the (noisy) starting sample.
\begin{equation*}
\begin{aligned}&\min _v\int_0^1\left\{\int_{\mathbb{R}^D \times \mathbb{R}^D}\left\|\left(\mathbf{x}_1-\mathbf{x}_0\right)-v\left(\mathbf{x}_t,t\right)\right\|^2 \pi\left(\mathbf{x}_0, \mathbf{x}_1\right) d \mathbf{x}_0 d \mathbf{x}_1\right\} d t \quad \text { //Flow Matching objective}\\= & \min _v\int_0^1 E\left[\left\|\left(\mathbf{x}_1-\mathbf{x}_0\right)-v\left(\mathbf{x}_{t}, t\right)\right\|^2 d t\right.\quad \text{// Rectified Flow objective} \\ = & \min _v E \int_0^1\left\|\left(\mathbf{x}_1-\mathbf{x}_0\right)-v\left(\mathbf{x}_{t}, t\right)\right\|^2 d t \quad \text{//Fubini–Tonelli Theorem }\\ \geqslant & \min _v E\left\|\int_0^1\left(\mathbf{x}_1-\mathbf{x}_0\right)-v\left(\mathbf{x}_t, t\right) d t\right\|^2 \quad  \text {//Jensen's inequality }\\ = & \min _v E\left\|\int_0^1\left(\mathbf{x}_1-\mathbf{x}_0\right) d t-\int_0^1 v\left(\mathbf{x}_{t}, t\right) d t\right\|^2 \\ = & \min _G E\left\|\mathbf{x}_1-\mathbf{x}_0 - \frac{G\left(\mathbf{x}_1, \mathbf{x}_t, t\right)}{t}\right\|^2 \quad \text { // Integration Flow by Equation (36),(37) }\end{aligned}
\end{equation*}

Therefore, Integration Flow models the transformation via a single integrated function $G$, which is sufficient and effectively optimal for solving the Flow-Matching/Rectified-Flow objective.
   
\end{proof}



\end{document}